%% file: paper.tex
\documentclass{article}
\usepackage{ijcai25}

\usepackage{times}
\usepackage{soul}
\usepackage{url}
\usepackage[hidelinks]{hyperref}
\usepackage[utf8]{inputenc}
\usepackage[small]{caption}
\usepackage{graphicx}
\usepackage{amsmath}
\usepackage{amsthm}
\usepackage{booktabs,colortbl}
\usepackage{algorithm}
\usepackage{algorithmicx}
\usepackage[noend]{algpseudocode}

\usepackage{microtype}
\usepackage{multirow}
\usepackage[dvipsnames]{xcolor}
\usepackage{etoolbox}
\usepackage{amssymb}
\usepackage{latexsym}
\usepackage{array}
\usepackage{adjustbox}
\usepackage{fancybox}
\usepackage{subcaption}
\usepackage{xspace}
\usepackage{setspace}
\usepackage{stmaryrd}
\usepackage{enumitem}
\usepackage{ifthen}
\usepackage{verbatim}
\usepackage{titlesec}
\usepackage[capitalise,nameinlink]{cleveref}
\usepackage[mode=text]{siunitx}
\usepackage{xfrac}
\usepackage{relsize}

\usepackage{tikz}
\usepackage{forest}

\usetikzlibrary{arrows,decorations.pathmorphing,decorations.footprints,fadings,calc,trees,mindmap,shadows,decorations.text,patterns,positioning,shapes,matrix,fit}
\usetikzlibrary{arrows,shadows,backgrounds}
\usetikzlibrary{arrows.meta}
\usetikzlibrary{positioning,fit}
\usetikzlibrary{automata}
\usetikzlibrary{matrix}
\usetikzlibrary{shapes.symbols,shapes.misc,shapes.arrows,shapes.geometric}
\usetikzlibrary{matrix,chains,scopes,decorations.pathmorphing}

\usepackage[round]{natbib}

\input{defs}

\input{preamble}

\begin{document}

\maketitle

\input{abs}
\input{intro}

\input{prelim}

\input{drxp}

\input{ecxp}

\input{xps}

\input{res}

\input{conc}

\input{replbib}
\input{togbbl} 

\iftoggle{mkbbl}{
    \bibliographystyle{named}
    \bibliography{team,refs}
}{

\input{paper.bibl}
}

\end{document}

%% file: defs.tex
\newboolean{nonanon}               
\setboolean{nonanon}{false}       %
\newboolean{mkcompact}             
\setboolean{mkcompact}{true}       %
\newboolean{squeezeit}             
\setboolean{squeezeit}{true}       %
\newtheorem{proposition}{Proposition}
\newtheorem{definition}{Definition}

\newtheorem{example}{Example}
\newtheorem{remark}{Remark}
\newtheorem{claim}{Claim}

\crefname{prop}{Proposition}{Propositions}
\Crefname{prop}{Proposition}{Propositions}
\crefname{defn}{Definition}{Definitions}
\Crefname{defn}{Definition}{Definitions}
\crefname{cor}{Corollary}{Corollaries}
\Crefname{cor}{Corollary}{Corollaries}
\crefname{exmpl}{Example}{Examples}
\Crefname{exmpl}{Example}{Examples}
\Crefname{theorem}{Theorem}{Theorem}

\newtheoremstyle{nutstyle}
{3.0pt} 
{3.0pt} 
{} 
{} 
{\bfseries} 
{.} 
{.5em} 
{} 

\theoremstyle{nutstyle}

\newtheoremstyle{nuxstyle}
{3.0pt} 
{3.0pt} 
{} 
{} 
{\itshape} 
{.} 
{.5em} 
{} 

\theoremstyle{nuxstyle}

\newcommand{\todoF}[2]{}

\newcommand{\fml}[1]{{\mathcal{#1}}}

\newcommand{\tn}[1]{\textnormal{#1}}
\newcommand{\mbf}[1]{\ensuremath\mathbf{#1}}
\newcommand{\msf}[1]{\ensuremath\mathsf{#1}}
\newcommand{\mbb}[1]{\ensuremath\mathbb{#1}}

\newcommand{\tbf}[1]{\textbf{#1}}

\newcommand{\pnorm}[1]{\ensuremath{l}_{#1}}
\newcommand{\refd}{\ensuremath\epsilon}

\newcommand{\SAT}{\ensuremath\mathsf{SAT}}
\newcommand{\True}{\textbf{true}}
\newcommand{\False}{\textbf{false}}

\newcommand{\prtaxp}{\ensuremath\mathsf{reportAXp}}
\newcommand{\prtcxp}{\ensuremath\mathsf{reportCXp}}

\newcommand{\findcxpswift}{\ensuremath\msf{SwiftCXp}}

\newcommand{\findcxpdicho}{\ensuremath\msf{FindCXpDichotomic}}
\newcommand{\robt}{\ensuremath\msf{FindAdvEx}}
\newcommand{\findaxp}{\ensuremath\msf{FindAXp}}
\newcommand{\findcxp}{\ensuremath\msf{FindCXp}}

\newcommand{\outc}{\oper{hasAE}}
\newcommand{\TRUE}{\textbf{true}\xspace}
\newcommand{\FALSE}{\textbf{false}\xspace}

\newcommand{\swiftcxp}{SwiftCXp\xspace}

\definecolor{darkslategray}{rgb}{0.18, 0.31, 0.31} 
\definecolor{platinum}{rgb}{0.9, 0.89, 0.89} 
\definecolor{gray}{rgb}{.4,.4,.4}
\definecolor{midgrey}{rgb}{0.5,0.5,0.5}
\definecolor{middarkgrey}{rgb}{0.35,0.35,0.35}
\definecolor{darkgrey}{rgb}{0.3,0.3,0.3}
\definecolor{darkred}{rgb}{0.7,0.1,0.1}
\definecolor{midblue}{rgb}{0.2,0.2,0.7}
\definecolor{darkblue}{rgb}{0.1,0.1,0.5}
\definecolor{defseagreen}{cmyk}{0.69,0,0.50,0}

\newcommand{\jnoteF}[1]{}
\newcommand{\anoteF}[1]{}

\newcommand{\oper}[1]{\ensuremath\textnormal{\small{\textsf{#1}}}}

\newcounter{tableeqn}[table]

\DeclareMathOperator*{\limply}{\rightarrow}

\newcommand{\bigland}{\ensuremath\bigwedge}
\newcommand{\waxp}{\ensuremath\mathsf{WAXp}}
\newcommand{\wcxp}{\ensuremath\mathsf{WCXp}}
\newcommand{\axp}{\ensuremath\mathsf{AXp}}
\newcommand{\cxp}{\ensuremath\mathsf{CXp}}
\newcommand{\dsym}{\ensuremath\mathfrak{d}} 
\newcommand{\tdsym}{$\mathfrak{d}$} 
\newcommand{\ewaxp}{\ensuremath\dsym\;\!\!\mathsf{WAXp}}
\newcommand{\ewcxp}{\ensuremath\dsym\;\!\!\mathsf{WCXp}}
\newcommand{\eaxp}{\ensuremath\dsym\;\!\!\mathsf{AXp}}
\newcommand{\ecxp}{\ensuremath\dsym\mathsf{CXp}}
\newcommand{\aex}{\ensuremath\mathsf{AEx}}

\newcommand{\FFA}{\ensuremath\mathsf{FFA}}

\newcommand{\twaxp}{WAXp\xspace}

\newcommand{\taxp}{AXp\xspace}
\newcommand{\tcxp}{CXp\xspace}

\newcommand{\tewcxp}{{\tdsym}\;\!\!WCXp\xspace}
\newcommand{\teaxp}{{\tdsym}\;\!\!AXp\xspace}
\newcommand{\tecxp}{{\tdsym}CXp\xspace}
\newcommand{\taex}{AEx\xspace}
\newcommand{\teaex}{{\tdsym}AEx\xspace}
\newcommand{\mailtodomain}[1]{\href{mailto:#1@icrea.cat}{\texttt{\nolinkurl{#1}}}}

\titleformat{\paragraph}[runin]
{\normalfont\bfseries}{}{0pt}{}

\newcolumntype{L}[1]{>{\raggedright\let\newline\\\arraybackslash\hspace{0pt}}m{#1}}
\newcolumntype{C}[1]{>{\centering\let\newline\\\arraybackslash\hspace{0pt}}m{#1}}
\newcolumntype{R}[1]{>{\raggedleft\let\newline\\\arraybackslash\hspace{0pt}}m{#1}}

\makeatletter
\newcommand\nparagraph{%
  \@startsection{paragraph}
    {4}
    {\z@}
    {0.225ex \@plus0.225ex \@minus.125ex}
    {-1em}
    {\normalfont\normalsize\bfseries}%
}
\makeatother

\titlespacing{\paragraph}{%
  0pt}{
  0.5\baselineskip}{
  1em}



%% file: preamble.tex
\title{Efficient Contrastive Explanations on Demand\footnote{%
The main purpose of this CoRR report is to serve as a timestamp 
against the copycats.}}

\iftrue
\author{
Yacine Izza$^1$
\And
Joao Marques-Silva$^2$\\
\affiliations
$^1$CREATE \& NUS, Singapore\\
$^2$ICREA, University of Lleida, Spain\\
\emails
izza@comp.nus.edu.sg,
jpms@icrea.cat
}
\fi

%% file: abs.tex
\begin{abstract}
  Recent work revealed a tight connection between adversarial
  robustness and restricted forms of symbolic explanations, namely
  distance-based (formal) explanations.
  This connection is significant because it represents a first
  step towards making the computation of symbolic explanations as
  efficient as deciding the existence of adversarial examples,
  especially for highly complex machine learning (ML) models.
  However, a major performance bottleneck remains, because of the very
  large number of features that ML models may possess, in particular 
  for deep neural networks.
  This paper proposes novel algorithms to compute the so-called 
  contrastive 
  explanations for ML models with a large number of features, by
  leveraging on adversarial robustness.
  Furthermore, the paper also proposes novel algorithms for
  listing  explanations and finding smallest contrastive explanations.
  The experimental results demonstrate the performance gains achieved
  by the novel algorithms proposed in this paper.
\end{abstract}

%% file: intro.tex
\section{Introduction}
\label{sec:intro}

The remarkable progress achieved by machine learning (ML) is largely
explained by the advances made in neural networks over the last two
decades.%
\footnote{The importance of these advances is demonstrated by the
recent Turing Awards and Nobel Prizes awarded to authors of some
seminal works related with neural
networks~\cite{acm-turing-award-2018,bbc-turing-award-2018,bbc-physics-nobel,bbc-chemistry-nobel}.}
A downside of complex neural networks (NNs) and other machine learning
models is their lack of interpretability, i.e. the operation of the
neural networks cannot be fathomed by human decision makers.
However, the ability to understand the rationale behind decisions is a
cornerstone to develop trustworthy systems of artificial intelligence
(AI).
Motivated by the challenge of understanding ML models, the last decade
witnessed a growing interest in explainable AI (XAI).
More recently, the need for rigor motivated the development of 
symbolic (formal) XAI approaches~\citep{ms-rw22,darwiche-lics23,ms-isola24}, including recent
promising results in the symbolic explanation of neural
networks~\citep{hms-corr23,barrett-nips23,swiftxp-kr24,barrett-corr24}, based on computing so-called
distance-restricted explanations. 
(Existing efficient formal XAI methods on other ML classifier 
families include but not limited 
to~\citep{ims-ijcai21,iims-corr20,hiims-kr21,iisms-aaai22,iims-jair22,iincms-corr22,ihincms-ijar23,iisms-aaai24} 
on tree-based models, \citep{izza-ecai24} on Binarized NNs, 
\citep{ims-sat21,Ignatiev-aaai21,hms-ecai23} on decision rules, 
\citep{NBC-nips20,ims-workshop-kr23} on Naive Bayes classifiers, etc.)
While recent work in the rigorous explanation of neural networks
focused on rule-based explanations (which are also referred to 
as sufficient or abductive explanations), the importance of
contrastive (also referred to as counterfactual) explanations cannot
be overstated~\citep{miller-aij19}.
Furthermore, it is well-known that there can exist multiple
(contrastive or abductive) explanations for a given sample. As a
result, another challenge is the enumeration of contrastive and/or
abductive explanations, which can be as well exploited for 
computing feature importance 
scores~\citep{izza-aaai24b,Ignatiev-sat24b,lhams-corr24a,lhms-aaai25}.

This paper proposes solutions to two concrete problems not studied in
earlier works. Concretely, the paper devises solutions for (i) the
computation of distance-restricted contrastive explanations; and (ii) 
the enumeration of distance-restricted contrastive (and abductive)
explanations. In addition to solving these two main problems, the
paper also proposes a number of additional contributions, which can be
summarized as follows: 

\begin{enumerate}[label={(\arabic*)}]
     \item Use of dedicated dichotomic search
algorithm for computing one contrastive explanation;  
    \item Effective parallelization of the dichotomic search algorithm, that enables
significant performance gains;  
    \item Novel heuristic for reducing the
number of iterations of the proposed dichotomic search algorithms; 
    \item Novel abstract-refinement approach for enumerating
	 distance-restricted explanations and its variant to find 
	 optimal contrastive explanations; 
    \item Experimental results demonstrating the scalability of the proposed algorithms
    	on large NNs. 
\end{enumerate}

%% file: prelim.tex
\section{Background}
\label{sec:prelim}

\paragraph{Measures of distance.}
%
The distance between two vectors $\mbf{v}$ and $\mbf{u}$ is denoted by
$\lVert\mbf{v}-\mbf{u}\rVert$, and the actual definition depends on
the norm being considered.
Different norms $\pnorm{p}$ can be considered.
For $p\ge1$, the $p$-norm is defined as follows~\citep{horn-bk12}:
\begin{equation}
  \begin{array}{lcl}
    \lVert\mbf{x}\rVert_{p} & {:=} &
    \left(\sum\nolimits_{i=1}^{m}|x_i|^{p}\right)^{\sfrac{1}{p}}
  \end{array}
\end{equation}

Let $d_i=1$ if $x_i\not=0$, and let $d_i=0$ otherwise. Then, for
$p=0$, we define the 0-norm, $\pnorm{0}$, as
follows~\citep{robinson-bk03}:
\begin{equation}
  \begin{array}{lcl}
    \lVert\mbf{x}\rVert_{0} & {:=} &
    \sum\nolimits_{i=1}^{m}d_i
  \end{array}
\end{equation}

In general, for $p\ge1$, $\pnorm{p}$ denotes the Minkowski distance.
Well-known special cases include
the Manhattan distance $\pnorm{1}$,
the Euclidean distance $\pnorm{2}$, and
the Chebyshev distance $\pnorm{\infty}=\lim_{p\to\infty}l_p$.
$\pnorm{0}$ denotes the Hamming distance.
In the remainder of the paper, we use $p\in\mbb{N}_0$ (but we also
allow $p=\infty$ for the Chebyshev distance).
%

\paragraph{Classification problems.}
%
Classification problems are defined on a set of features
$\fml{F}=\{1,\ldots,m\}$ and a set of classes
$\fml{K}=\{c_1,\ldots,c_K\}$.
Each feature $i$ has a domain $\mbb{D}_i$. Features can be ordinal or
categorical. Ordinal features can be discrete or real-valued.
Feature space is defined by the cartesian product of the features'
domains: $\mbb{F}=\mbb{D}_1\times\cdots\times\mbb{D}_m$.
A classifier computes a total function $\kappa:\mbb{F}\to\fml{K}$.
Throughout the paper, a classification problem $\fml{M}$ represents a
tuple $\fml{M}=(\fml{F},\mbb{F},\fml{K},\kappa)$.

An instance (or a sample) is a pair $(\mbf{v},c)$, with
$\mbf{v}\in\mbb{F}$ and $c\in\fml{K}$.
An explanation problem $\fml{E}$ is a tuple
$\fml{E}=(\fml{M},(\mbf{v},c))$. The generic purpose of XAI is to find
explanations for each given instance.
Moreover, when reasoning in terms of robustness, we are also
interested in the behavior of a classifier given some instances.
Hence, we will also use explanation problems when addressing
robustness.

\paragraph{Running example.}
To illustrate some of the definitions in this section, the following 
very simple classifiers are used as the running examples 
throughout the paper.

\begin{example} \label{ex:runex}
  Let us consider the following classification
  problem.
  The features are $\fml{F}=\{1,2,3\}$, all ordinal with domains
  $\mbb{D}_1=\mbb{D}_2=\mbb{D}_3=\mbb{R}$. The set of classes is
  $\fml{K}=\{0,1\}$. Finally, the classification function is
  $\kappa:\mbb{F}\to\fml{K}$, defined as follows (with
  $\mbf{x}=(x_1,x_2,x_3)$): 
  \[
  \kappa(\mbf{x})=\left\{
  \begin{array}{lcl}
    1 & ~~ & \tn{if~} 0<{x_1}<2 \land 4x_1\ge(x_2+x_3) \\[3pt]
    0 & ~~ & \tn{otherwise}
  \end{array}
  \right.
  \]
  Moreover, let the target instance be $(\mbf{v},c)=((1,1,1),1)$.
\end{example}

\paragraph{Adversarial examples.}
%
Let $\fml{M}=(\fml{F},\mbb{F},\fml{K},\kappa)$ be a classification
problem. Let $(\mbf{v},c)$, with $\mbf{v}\in\mbb{F}$ and
$c=\kappa(\mbf{v})$, be a given instance. Finally, let $\refd>0$ be
a value of distance for norm $l_p$.

We say that there exists an adversarial example if the following logic
statement holds true,
\begin{equation} \label{eq:locrob}
  \exists(\mbf{x}\in\mbb{F}).\left(\lVert\mbf{x}-\mbf{v}\rVert_p\le\refd\right)\land\left(\kappa(\mbf{x})\not=c\right) 
\end{equation}
(The logic statement above holds true if there exists a point
$\mbf{x}$ which is less than $\refd$ distance (using norm $\pnorm{p}$)
from $\mbf{v}$, and such that the prediction changes.)
If~\eqref{eq:locrob} is false, then the classifier is said to be
$\refd$-robust.
If~\eqref{eq:locrob} is true, then any $\mbf{x}\in\mbb{F}$ for which
the following predicate holds:\footnote{Parameterizations are shown as
predicate arguments positioned after ';'. These may be dropped for the
sake of brevity.}
%
\begin{equation} \label{eq:ae}
  \aex(\mbf{x};\fml{E},\refd,p) ~:=~
  \left(\lVert\mbf{x}-\mbf{v}\rVert_p\le\refd\right)\land\left(\kappa(\mbf{x})\not=c\right) 
\end{equation}
is referred to as an \emph{adversarial example}.%
Tools that decide the existence of adversarial examples will be
referred to as \emph{robustness tools}. (In the case of neural
networks (NNs), the progress observed in robustness tools is
documented by VNN COMP~\citep{johnson-sttt23}.)

It may happen that we are only interested in inputs that respect some
constraint, i.e.\ not all points of feature space are allowed or 
interesting. In such cases, we define adversarial examples subject to
some constraint $\fml{C}:\mbb{F}\to\{0,1\}$, which are referred to as
\emph{constrained AExs}. In this case, the adversarial examples must
satisfy the following logic statement:
\begin{equation} \label{eq:ae2}
  \fml{C}(\mbf{x})\land\left(\lVert\mbf{x}-\mbf{v}\rVert_p\le\refd\right)\land\left(\kappa(\mbf{x})\not=c\right)
\end{equation}
Clearly, the predicate $\aex$ (see~\eqref{eq:ae}) can be parameterized
by the constraint being used.


\begin{example} \label{ex:runex:ae}
  For the classifier from~\cref{ex:runex}, for distance $l_1$, and
  with $\refd=1$, there exist adversarial examples by either setting
  $x_1=0$ or $x_1=2$.
\end{example}

\input{sxp}

%% file: sxp.tex
\paragraph{Symbolic explanations.}
%
This work builds on rigorous, symbolic (or logic-based)
explanations introduced in earlier 
work~\citep{darwiche-ijcai18,inms-aaai19}. 
Mainly, symbolic explanations are conventionally categorized into 
two types: 
abductive~\citep{darwiche-ijcai18,inms-aaai19} and
contrastive~\citep{miller-aij19,inams-aiia20}.
%
Abductive explanations (AXps) broadly answer a \tbf{Why} question,
i.e.\ \emph{Why the prediction?}, whereas contrastive explanations
(CXps) broadly answer a \tbf{Why Not} question, i.e.\ \emph{Why not
some other prediction?}.
Intuitively, an AXp is a subset-minimal set of feature values
$(x_i=v_i)$, at most one for each feature $i\in\fml{F}$, that is
sufficient to trigger a particular class and satisfy the instance
being explained.
Similarly, a CXp is a subset-minimal set of features by changing the
values of which one can trigger a class different from the target one.
More formally, AXps and CXps are defined below.

Given an explanation problem $\fml{E}=(\fml{M}, (\mbf{v}, c))$, an
\emph{abductive explanation (AXp) of $\fml{E}$} is a subset-minimal set 
$\fml{X}\subseteq\fml{F}$ of features which, if assigned the values dictated by
the instance $(\mbf{v},c)$, are sufficient for the prediction.
The latter condition is formally stated, for a set $\fml{X}$, as follows:
\begin{equation} \label{eq:axp}
\forall(\mbf{x}\in\mbb{F}).\left[\bigland\nolimits_{i\in\fml{X}}(x_i=v_i)\right]\limply(\kappa(\mbf{x})=c)
\end{equation}
Any subset $\fml{X} \subseteq \fml{F}$ that satisfies \eqref{eq:axp},
but is not subset-minimal (i.e.\ there exists $\fml{X}^\prime \subset
\fml{X}$ that satisfies~\eqref{eq:axp}), is referred to as a
\emph{weak abductive explanation (Weak AXp)}.

\begin{example}
  For the classifier from~\cref{ex:runex}, 
  the AXp of the instance $\mbf{v}$ is $\{1, 2, 3\}$.
\end{example}

Given an explanation problem, a contrastive explanation
(CXp) is a subset-minimal set of features $\fml{Y}\subseteq\fml{F}$
which, if the features in $\fml{F}\setminus\fml{Y}$ are assigned the
values dictated by the instance $(\mbf{v},c)$, then there is an
assignment to the features in $\fml{Y}$ that changes the prediction.
This is stated as follows, for a chosen set $\fml{Y}\subseteq\fml{F}$:
\begin{equation} \label{eq:cxp}
  \exists(\mbf{x}\in\mbb{F}).\left[\bigland\nolimits_{i\in\fml{F}\setminus\fml{Y}}(x_i=v_i)\right]\land(\kappa(\mbf{x})\not=c)
\end{equation}

AXp's and CXp's respect a minimal-hitting set (MHS) duality
relationship~\cite{inams-aiia20}.
Concretely, each AXp is an MHS of the CXp's and each CXp is an MHS of
the AXp's.
MHS duality is a stepping stone for the enumeration of explanations.

Furthermore, recent work on XAI \citep{msm-ijcai20} relates the concept abductive and 
contrastive explanations in the context of ML model explanability to, respectively, 
minimal unsatisfiable subsets (MUSes) and minimal correction subsets (MCSes) 
concepts in the context of logic formulas~\citep{sat-handbook21,msjm-aij17}. 
Clearly advances in MUSes/MCSes computation can be easily adapted 
for computing symbolic explanations.

%% file: drxp.tex
Next, we present a generalized definition of (W)AXps and (W)CXps,
that take into account the $\pnorm{p}$ distance between $\mbf{v}$ and
the points that can be considered in terms of changing the prediction
$c=\kappa(\mbf{v})$. 
First, we overview the formal definition of distance-restricted AXps/CXps,
i.e.\ {\teaxp}a/{\tecxp}s proposed in~\cite{swiftxp-kr24}, which take the $\pnorm{p}$ distance into
account. Afterwards, we show a number of properties
on {\teaxp}a/{\tecxp}s, 
including the MHS duality between the two explanation classes.

\paragraph{Distance-restricted AXps/CXps.} 
\label{sec:drxp}
%
The standard definitions of AXps \& CXps can be generalized to take a
measure $\pnorm{p}$ of distance into account.

\begin{definition}[Distance-restricted (W)AXp, \tdsym(W)AXp]
  For a norm $\pnorm{p}$, a set of features $\fml{X}\subseteq\fml{F}$
  is a distance-restricted weak abductive explanation (\twaxp) for an
  instance $(\mbf{v},c)$, within distance $\refd>0$ of $\mbf{v}$, if
  the following predicate holds true,
  \begin{align} \label{eq:waxpg}
    &\ewaxp(\fml{X};\fml{E},\refd,p) ~:=~ \forall(\mbf{x}\in\mbb{F}).\\
    & \left(\bigwedge\nolimits_{i\in\fml{X}}(x_i=v_i)\land(\lVert\mbf{x}-\mbf{v}\rVert_{p}\le\refd)\right)\limply(\kappa(\mbf{x})=c)\nonumber
  \end{align}
  If a (distance-restricted) weak AXp $\fml{X}$ is irreducible
  (i.e.\ it is subset-minimal), then $\fml{X}$ is a
  (distance-restricted) AXp, or \teaxp.
\end{definition}

\begin{definition}[Distance-restricted (W)CXp, \tdsym(W)CXp]
  For a norm $\pnorm{p}$, a set of features $\fml{Y}\subseteq\fml{F}$
  is a weak contrastive explanation (WCXp) for an instance
  $(\mbf{v},c)$, within distance $\refd>0$ of $\mbf{v}$, if the
  following predicate holds true,
  \begin{align} \label{eq:wcxpg}
    &\ewcxp(\fml{Y};\fml{E},\refd,p) ~:=~ \exists(\mbf{x}\in\mbb{F}). \\
    &\left(\bigwedge\nolimits_{i\in\fml{F}\setminus\fml{Y}}(x_i=v_i)\land(\lVert\mbf{x}-\mbf{v}\rVert_{p}\le\refd)\right)\land(\kappa(\mbf{x})\not=c) \nonumber
  \end{align}
  If a (distance-restricted) weak CXp $\fml{Y}$ is irreducible,
  then $\fml{Y}$ is a (distance-restricted) CXp, or \tecxp.
\end{definition}

Furthermore, when referring to {\teaxp}s (resp.~{\tecxp}s), the
predicates $\eaxp$ (resp.~$\ecxp$) will be used.

\begin{example}
  For the classifier of~\cref{ex:runex}, let the norm used be $l_1$,
  with distance value $\refd=1$. From~\cref{ex:runex:ae}, we know that
  there exist adversarial examples, e.g.\ by setting $x_1=0$ or
  $x_1=2$. However, if we fix the value of $x_1$ to 1, then any
  assignment to $x_2$ and $x_3$ with $|x_2-1|+|x_3-1|\le1$, will not
  change the prediction. As a result, $\fml{X}=\{1\}$ is a
  distance-restricted AXp when $\refd=1$. Moreover, by allowing only
  feature 1 to change value, we are able to change prediction, since
  we know there exists an adversarial example.
\end{example}

\begin{remark}
  Distance unrestricted AXps (resp.~CXps) correspond to
  $m$-distance {\teaxp}s (resp.~{\tecxp}s) for norm $l_0$, where $m$
  is the number of features.
\end{remark}

\paragraph{Relating \tdsym(W)CXps \& {\taex}s.}
An important observation to underscore is the preserved connection
between adversarial examples and weak \teaxp. Namely, 
if there exists an adversarial example then it must yield to 
a (weak) \tecxp. 
Similarly, the existence of a (weak) \tecxp implies there exists 
an adversarial example and it is consistent with the \tecxp.

\begin{proposition} \label{prop:aex2xp}
  Consider an explanation problem $\fml{E}=(\fml{M},(\mbf{v},c))$ and
  some $\refd>0$ for norm $\pnorm{p}$. Let $\mbf{x}\in\mbb{F}$, with
  $\lVert\mbf{x}-\mbf{v}\rVert_p\le\refd$, and let
  $\fml{D}=\{i\in\fml{F}\,|\,x_i\not=v_i\}$. 
  Then,
  \begin{enumerate}[nosep]
  \item If $\aex(\mbf{x};\fml{E},\refd,p)$ holds, then
    $\ewcxp(\fml{D};\fml{E},\refd,p)$
    holds;
  \item If $\ewcxp(\fml{D};\fml{E},\refd,p)$
    holds, then
    $\exists(\mbf{y}\in\mbb{F}).
    \lVert\mbf{y}-\mbf{v}||_p\le||\mbf{x}-\mbf{v}\rVert_p\land
    \aex(\mbf{y};\fml{E},\refd,p)$.
  \end{enumerate}
\end{proposition}

\paragraph{Distance-restricted {\taxp}s/{\tcxp}s duality.}
Given the definitions above, we define the set of all {\teaxp}'s
and set of all {\teaxp}'s as follows:
\begin{align}
  \dsym\mbb{A}(\fml{E},\refd;p) &= \{\fml{X}\subseteq\fml{F}\,|\,\eaxp(\fml{X};\fml{E},\refd,p)\}
  \label{eq:allaxpg}\\
  \dsym\mbb{C}(\fml{E},\refd;p) &= \{\fml{Y}\subseteq\fml{F}\,|\,\ecxp(\fml{Y};\fml{E},\refd,p)\}
  \label{eq:allcxpg} 
\end{align}


In turn, this yields the following result regarding MHS
 duality between 
 {\teaxp}s \& {\tecxp}s.

\begin{proposition} \label{prop:duality2}
  Given an explanation problem $\fml{E}$, norm $\pnorm{p}$, and a
  value of distance $\refd>0$ then,
  \begin{enumerate}[nosep]
  \item A set $\fml{X}\subseteq\fml{F}$ is a \teaxp iff $\fml{X}$ is a
    MHS of the {\tecxp}s in
    $\dsym\mbb{C}(\fml{E},\refd;p)$.
  \item A set $\fml{Y}\subseteq\fml{F}$ is a \tecxp iff $\fml{Y}$ is a
    MHS of the {\teaxp}s in
    $\dsym\mbb{A}(\fml{E},\refd;p)$.
  \end{enumerate}
\end{proposition}

MHS duality between %
 $\dsym\mbb{A}(\fml{E},\refd;p)$ and $\dsym\mbb{C}(\fml{E},\refd;p)$ 
 exhibits a special case when there are no adversarial examples, i.e. 
 when the predicate $\aex(\mbf{x};\fml{E},\refd,p)$ (see \eqref{eq:ae}) 
 does not hold for any $\mbf{x}\in \mbb{F}$.\footnote{%
 For example, when the given $\refd$ is so small that 
 the ML model is constant within the considered $\refd$-ball.}
In such a case, there is no \tecxp and the unique \teaxp is the empty
set.
As can be concluded, \cref{prop:duality2} holds even in such a
situation.

\cref{prop:duality2} is instrumental for the enumeration of {\teaxp}s
and {\tecxp}s, as shown in earlier work in the case of
distance-unrestricted AXps/CXps~\citep{inams-aiia20}, since it enables
adapting well-known algorithms for the enumeration of subset-minimal
reasons of inconsistency~\citep{lpmms-cj16}.

\begin{example}
  For the running example, we have that
  $\dsym\mbb{A}(\fml{E},1;1)=\dsym\mbb{C}(\fml{E},1;1)=\{\{1\}\}$.
\end{example}

The definitions of distance-restricted CXps also reveal
novel uses for contrastive explanations.
For a given distance $\refd$ and a point $\mbf{v}$, smallest 
(minimum) \teaxp represents 
a sufficient reason to demonstrate that the classifier is locally 
(in the vicinity of $\mbf{v}$) not robust, i.e.\ it has an adversarial
example.

%% file: ecxp.tex
\section{Computing Contrastive Explanations}
\label{sec:ecxp}

As aforementioned earlier,  algorithms for computing CXps
build on those for computing MCSes of logic formulas. The same
observation can be made in the case of distance-restricted CXps.
%

\subsection{Baseline Algorithms for Computing  {\tecxp}s}
%
Throughout this section, we assume that the existence of adversarial 
examples is decided by calls to a suitable oracle. In the algorithms 
described in this section, this oracle is represented by a predicate 
$\robt$. 
Furthermore, we will
require that the robustness oracle allows some features to be fixed,
i.e.\ the robustness oracle decides the existence of constrained
adversarial examples. As a result, the call to $\robt$ uses as
arguments the distance $\epsilon$ and the set of fixed features, and
it is parameterized by the explanation problem $\fml{E}$ and the norm $p$
used.

Recall that in cases where no AEx exists in the $\refd$ distance 
neighborhood, then no \tecxp is reported  and \teaxp = $\emptyset$. 
Clearly, the algorithms discussed above for finding one \tecxp would 
require first instrumenting an oracle call 
to verify that there exists at least one \tewcxp (i.e.\  
$\robt(\epsilon,\emptyset;\fml{M},(\mbf{v},c),p) = \True$); in the negative 
case the algorithm prints that $\refd$ is too small to include 
a \tecxp and terminates. 
Moreover, we will show later how this initial call serves to generate 
an approximation of \tecxp (\tewcxp), i.e.\  the initial feature 
set $\fml{W}\subseteq\fml{F}$ to inspect.

\paragraph{Transition feature.}
One key concept in \tcxp extraction is identifying
the \emph{transition features}. 
Given some set $\fml{X} = \fml{F}\setminus\fml{S}$ of fixed features,
$i\in\fml{F}$ is a transition feature if (a) when $i$ is not fixed
(i.e.\ $i\in\fml{S}$), then an adversarial example exists;
and (b) when $i$ is fixed (i.e.\ $i\not\in\fml{S}$), then no adversarial
example exists. The point is that $i$ must be included in $\fml{S}$
for $\fml{S}$ to represent a \tewcxp.

\paragraph{Dichotomic search algorithm.}
%
Aiming to reduce the overall running time of computing one \teaxp,
this paper seeks mechanisms to avoid the $\Theta(|\fml{F}|)$
\emph{sequential} calls to the robustness oracle. For that, it will be
convenient to study another (less used) algorithm, one that implements
dichotomic (or binary) search~\citep{sais-ecai06}.
\begin{algorithm}[t]
  \input{./algs/findcxp_dicho}
  \caption{Dichotomic search to find one \tecxp}
  \label{alg:dicho}
\end{algorithm}
The dichotomic search algorithm is shown in~\cref{alg:dicho}.%
\footnote{%
With a slight abuse of notation, set $\fml{W}$ is assumed to be
ordered, such that $\fml{W}_{a..b}$ denotes picking the elements
(i.e.\ features) ordered from index $a$ up to $b$. It is also assumed
that $\fml{W}_{a..b}$, with $a=0\lor{a>b}$ represents an empty set.}
At each iteration of the outer loop, the algorithm uses binary
search in an internal loop to find a transition feature,
i.e.\ freeing the features in $\fml{S}\cup\fml{W}_{1..j}$ 
yield an \taex, but freeing only the features in
$\fml{S}\cup\fml{W}_{1..i-1}$ does not exhibits an \taex, if
$i\not=0$.
(Upon termination of the inner loop, if $i=0$, then
$\fml{W}=\emptyset$.)
Moreover, the features in $\fml{W}_{j+1..{|\fml{W}|}}$ can be safely
discarded.
As a result, it is the case that the inner loop of the algorithm
maintains the following two invariants: (i)
$\ewcxp(\fml{S}\cup\fml{W}_{1..j})$; and (ii)
$(i=0)\lor\neg\ewcxp(\fml{S}\cup\fml{W}_{1..i-1})$. 
%
Clearly, the updates to $i$ and $j$ in the inner loop maintain the
invariants.
Moreover, it is easy to see that the features in $\fml{S}$ denote a
subset of a \tecxp, since we only add to $\fml{S}$ transition
features; this represents the invariant of the outer loop. When there
are no more features to analyze, then $\fml{S}$ will denote a
\tecxp.
If $k_M$ is the size of the largest \tecxp, then the number of calls
to the robustness oracle is $\fml{O}(k_M\log{m})$.
If the largest \tecxp is significantly smaller than $\fml{F}$, then
one can expect dichotomic search to improve the performance 
w.r.t the linear search algorithm presented next.

Our intuition is to vision the dichotomic search algorithm as a
procedure for analyzing chunks of features. 
We will later see that parallelization can be elicited by analyzing 
different chunks of features in parallel.

\paragraph{Basic linear search algorithm.}
%
It is worth noting that the linear search and its variants has been largely 
applied for computing {\taxp}'s and more recently for 
{\teaxp}'s  \citep{barrett-corr22,hms-corr23,barrett-nips23}.
Since {\teaxp}s and {\tecxp}s are also examples of
MSMP (minimal sets over a monotone
predicate~\cite{msjb-cav13,msjm-aij17}), then the same algorithm can
also be used for computing one \tecxp.
%
%
%
Roughly, the construction of the \tecxp $\fml{S}$ is achieved by a greedy/linear 
inspection over all input features while maintaining \cref{eq:wcxpg} 
hold true for $\fml{S}$. More precisely, the  algorithm iteratively picks 
a feature to be allowed to be
unconstrained, starting by fixing all features to the values dictated
by $\mbf{v}$. If no adversarial example is identified, then the
feature is left unconstrained; otherwise, it becomes fixed again.
%
%
Clearly, the algorithm requires $\Theta(|\fml{F}|)$ calls to
the robustness oracle. 

\paragraph{CLD algorithm.}
The \emph{clause} $D$ (CLD) algorithm~\citep{mshjpb-ijcai13}  
is designed to compute MCS (\emph{minimal correction set}) for 
over-constrained problems,  
where elements that can be dropped from the minimal set are
iteratively identified, and several can be removed in each oracle
call.
In contrast with algorithms for MUSes and AXps, an MCS (or CXp) can be
decided with a single call using the so-called clause $D$ (or
disjunction clause).
If the elements represented in the clause $D$ represent a minimal set,
then no additional elements can be found, and so the algorithm
terminates by reporting the minimal set~\citep{mshjpb-ijcai13}.
As shown later in the paper, we can use parallelization to emulate the
CLD algorithm in the case of computing one \tecxp,

\paragraph{Discussion.}
The algorithms outlined in this section, or the examples used in
recent work~\citep{barrett-corr22,hms-corr23,barrett-nips23} for 
computing {\teaxp}s, link the
performance of computing explanations to the ability of deciding the
existence of adversarial examples. More efficient tools for deciding
the existence of adversarial examples
(e.g.\ from~\cite{johnson-sttt23}) will result in more efficient
algorithms for computing distance-restricted explanations.
Nevertheless, one bottleneck of the algorithms discussed in this
section is that the number of calls to an oracle deciding the
existence of an adversarial example grows with the number of features.
For complex ML models with a large number of features, the overall
running time can become prohibitive.
The next section outlines novel insights on how to reduce the overall
running time by exploiting opportunities to parallelize calls to the
robustness oracle.

\begin{algorithm}[ht]
\input{./algs/swiftcxp}

  \caption{\swiftcxp algorithm to find one \tecxp}
  \label{alg:swiftxp}
\end{algorithm}

\subsection{\swiftcxp Algorithm}
The main intuition of our new algorithm, called \swiftcxp,  
is to implement a dichotomic search 
for analyzing (possibly in parallel) different chunks of features. 
%
Moreover,  we integrate in \swiftcxp a parallelized variant 
of the CLD technique which enables improving significantly 
its performances in practice.


\cref{alg:swiftxp} outlines the \swiftcxp algorithm, which runs in
parallel on multi-core CPU or GPU.
The procedure takes as input the explanation problem $\fml{E}$, an
$l_p$ distance $\epsilon > 0$, a threshold $\delta \in [0, 1]$ used
for activating the optional feature disjunction check, and the number
$q$ of available processors; and returns a \teaxp
$\fml{S}\subseteq\fml{F}$.
Intuitively, the algorithm implements a parallel dichotomic search by
splitting the set of features to analyze into a collection of chunks
and instruments decision oracle calls checking the existence of
adversarial examples, done in parallel on those chunks.
Upon completion of such a parallel oracle call, the algorithm proceeds
by zooming into a chunk that is deemed to contain a transition
feature.

The algorithm starts by initializing the operational set of features
$\fml{W}$ to contain all the features of $\fml{F}$ and a
subset-minimal \tecxp $\fml{S}$ to extract as $\emptyset$.
Note that one can potentially impose a heuristic feature order on
$\fml{F}$ aiming to quickly remove irrelevant features.
Additionally, one can compute an approximation of the initial 
subset features $\fml{W}$ to inspect rather than starting  
with the entire feature set $\fml{F}$.  
As we will detail it in the evaluation section, this can be
achieved by a single (initial) call to the robustness oracle 
and identify  the subset features where the values are 
flipped/changed in the \teaex.
In each iteration of the outer loop, the lower and upper bounds $\ell$
and $u$ on feature indices are set, respectively, to 1 and $|\fml{W}|$.
\begin{algorithm}[ht]

\input{./algs/featD}
  \caption{Parallelized feature disjunction check}
  \label{alg:fd}
\end{algorithm}

Each transition element is determined in parallel by the inner loop of
the algorithm, which implements dichotomic search and iterates until
$\ell+1 = u$.
An iteration of this loop splits the set of features $\fml{W}$ into
$\omega$ chunks determined by the splitting indices kept in
$\fml{D}\subseteq\fml{W}$.
(Note that the value of $\omega$ equals either the number of available
CPUs $q$ or the number of remaining features in $\fml{W}$, depending
on which of these values is smaller.)
Given the largest feature index $i\in\fml{D}$ in each such chunk, the
iteration tests whether an adversarial example can be found while
fixing the features $\fml{S}\cup\fml{W}_{1..i}$.
The test is applied in parallel for all the splitting indices
$i\in\fml{D}$ employing $\omega$ CPUs.

The aim of the algorithm is to determine the first case when an oracle
call reports that an adversarial example exists, i.e. that
$\oper{AE}_t=\True$ s.t. $\oper{AE}_{\ge t}=\True$ and
$\oper{AE}_{<t}=\False$.
Importantly, as soon as such case $t$ is determined, all the parallel
jobs are terminated.
(We underline that in practice terminating the jobs after $i$ s.t.
$\oper{AE}_i=\True$ and before $i$ s.t. $\oper{AE}_i=\False$ helps to
save a significant amount of time spent on the parallel oracle calls.)
The inner loop proceeds by zooming into the $t$'th chunk of features
by updating the values of the lower and upper bounds $\ell$ and $u$ as
it is deemed to contain a transition feature.
If all the oracle calls unanimously decide that an adversarial example
exists (resp.\  does not exist), the algorithm proceeds by zooming into the
corresponding \emph{boundary} chunk of features 
(i.e.\ $\ell$, resp.\ $u$).
Note that if the value of the upper bound $u$ is updated from
$|\fml{W}|$ all the way down to 1, which happens if all the parallel
oracle calls report an adversarial example, the algorithm needs to
check whether set $\fml{S}$ is sufficient for the given prediction.
If it is the case, the algorithm terminates by reporting $\fml{S}$.
Otherwise, it collects a newly determined transition feature and
proceeds by updating $\fml{W}$ and $\fml{S}$.


Ultimately, we devise an analogue of the CLD procedure widely 
used in the computation of MCS of an unsatisfiable 
logic formula~\citep{mshjpb-ijcai13}.
The analogue is referred to as feature disjunction check (see
$\msf{FeatDisjunct}$ in \cref{alg:fd}) and used as an optional
optimization step in \cref{alg:swiftxp} at the beginning of the main
(outer) loop.
We implement a heuristic order over $\fml{F}$ and activate
$\msf{featureDisjunct}$ (given some threshold $\delta$) at the last
iterations of $\findcxpswift$, where it is likely to conclude that all
the features in a selected subset $\fml{T}\subseteq\fml{W}$ of size
$\min(q, |\fml{W}|)$ are relevant for the explanation and can be
safely moved to $\fml{S}$ at once; otherwise, one can randomly pick a
single feature in $\fml{T}$ among those verified as irrelevant
features, i.e.\ removing the feature yields an adversarial
example, and fix it.
Finally, we observe that after running $\msf{featureDisjunct}$, the
algorithm does not invoke dichotomic search in the subsequent
iterations.

%% file: algs/findcxp_dicho.tex
\begin{flushleft}
  \hspace*{\algorithmicindent}
  \textbf{Input}: {
    Arguments: 
    $\epsilon$;
    Parameters: 
    $\fml{E}$,
    $p$}\\
  \hspace*{\algorithmicindent}
  \textbf{Output}: {One \tecxp $\fml{S}$}
\end{flushleft}

\begin{algorithmic}[1]
  \Function{$\findcxpdicho$}{$\epsilon;\fml{E},p$}
  \State{$(\fml{S},\fml{W})\gets(\emptyset,\fml{F})$} 
  \Comment{\footnotesize{Precondition: $\ewcxp(\fml{S}\cup\fml{W})$}}
  \While{$\fml{W}\not=\emptyset$}
  \Comment{\footnotesize{Invariant: $\exists(\fml{X}\in\dsym\mbb{C}).\fml{S}\subseteq\fml{X}$}}
  \State{$(i,j)\gets(0,|\fml{W}|)$}
  \While{$i<j$} 
  \Comment{\footnotesize{Invariant $\ewcxp(\fml{S}\cup\fml{W}_{1..j})$}}
  \State{$t\gets\lfloor\sfrac{(i+j)}{2}\rfloor$}
  \State{$\outc=\robt(\epsilon,\fml{F}\setminus\fml{S}\cup\fml{W}_{1..t};\fml{E},p)$}
  \If{$\outc$} 
  \State{$j\gets{t}$}
  \Comment{Fix more features}  
  \Else
  \State{$i\gets{t+1}$}
  \Comment{Free more features}
  \EndIf
  \EndWhile
  \State{$(\fml{S},\fml{W})\gets(\fml{S}\cup\fml{W}_{j..j},\fml{W}_{1..j-1})$}
  \EndWhile
  \State{\tbf{return} $\fml{S}$}
  \Comment{\footnotesize{$\exists(\fml{X}\:\!{\in}\:\!\dsym\mbb{C}).(\fml{S}\:\!{\subseteq}\:\!\fml{X})\:\!{\land}\:\!(\fml{W}\:\!{=}\:\!\emptyset)\:\!{\limply}\:\!\ecxp(\fml{S})$}}
  \EndFunction
\end{algorithmic}

%% file: algs/swiftcxp.tex
\begin{flushleft}
  \hspace*{\algorithmicindent}
  \textbf{Input}: {
    Arguments:   $\epsilon$, $q$, $\delta$;
    Parameters: $\fml{E}$,  $p$}\\
\hspace*{\algorithmicindent}
\textbf{Output}: {One \tecxp $\fml{S}$}
\end{flushleft}
\begin{algorithmic}[1]
  \Function{$\findcxpswift$}{$\epsilon, q, \delta;\fml{E},p$}
  \State{$(\fml{W}, \fml{S}) \gets (\fml{F},  \emptyset)$}
  \Comment{\footnotesize{Precond: $\ewcxp(\fml{F})\land(q\ge2)$}}
  \While{$\fml{W} \neq \emptyset$}
       \Comment{\footnotesize{\footnotesize{$\fml{S}\subseteq\fml{X}\in\dsym\mbb{C}$}}}
       \If{ $|\fml{W}| < \delta\times |\fml{F}|$ }
       	  \Comment{\footnotesize{Run FD check}}
          \State{$(\fml{W},\fml{S})\gets\msf{FeatDisjunct}(\epsilon, q, \fml{W}, \fml{S}; \fml{E},p)$}
          \State{\bfseries{continue}}
       \EndIf
      \State{$(\ell,u)\gets(0,|\fml{W}|)$}
      \While{$\ell+1<u$}
           \Comment{\footnotesize{Inv. $\ewcxp(\fml{S}\cup\fml{W}_{1..u})$}}
           \State{$\omega\gets\min(q, u-\ell)$}
           \Comment{\# parallel calls}
           \State{$\sigma\gets\lfloor\sfrac{(u-\ell)}{\omega}\rfloor$}
           \Comment{$\sigma$: chunk size}
           \State{$\fml{D}\gets\left\{\ell+\iota\times\sigma\;|\;\iota\in\{1,\ldots,\omega\}\right\}$}%
           \algrenewcommand\algorithmicdo{\textbf{do in parallel}}
           \For{$i \in \fml{D}$}
           \State{$\oper{AE}_i{\gets}\robt(\epsilon,\fml{F}\setminus\fml{S}\;\!{\cup}\;\!\fml{W}_{1..i};\fml{E},p)$}
           \EndFor
           \algrenewcommand\algorithmicdo{\textbf{do}}
           \State{$u\gets\min(\{i\in\fml{D}\mid\oper{AE}_i=\False\}\cup\{u\})$}
           \State{$\ell\gets\max(\{i\in\fml{D}\mid i<u\}\cup\{\ell\})$}        
       \EndWhile
       \If{${u=1}\land\robt(\epsilon,\fml{F}\setminus\fml{S};\fml{E},p)$} %
       \State{\Return{$\fml{S}$}}
       \EndIf
       \State{$(\fml{W}, \fml{S}) \gets (\fml{W}_{1..u-1}, \fml{S}\cup\fml{W}_{u..u})$}
       \EndWhile
  \State\Return{$\fml{S}$}
\EndFunction
\end{algorithmic}

%% file: algs/featD.tex
\begin{flushleft}
  \hspace*{\algorithmicindent}
  \textbf{Input}: {
    Arguments:   $\epsilon$, $\fml{W}$, $\fml{S}$;
    Parameters: $\fml{E}$,  $p$}\\
\end{flushleft}
\begin{algorithmic}[1]
  \Procedure{$\msf{FeatDisjunct}$}{$\epsilon, q, \fml{W}, \fml{S}; \fml{E}, p$}
  \State{$\fml{T}\gets\fml{W}_{\max(|\fml{W}|-q+1,1)..|\fml{W}|}$}
           \algrenewcommand\algorithmicdo{\textbf{do in parallel}}
           \For{$i \in \fml{T} $}
               \State{$\oper{AE}_i\gets\robt(\epsilon,\fml{F}\setminus(\fml{S}\cup\fml{W}\setminus\{i\});\fml{E},p)$}
           \EndFor
           \algrenewcommand\algorithmicdo{\textbf{do}}
       \If{$\bigland\nolimits_{i \in \fml{T}} \oper{AE}_i = \False$}
           \Comment{Free more features}
           \State\Return{$(\fml{W}\setminus \fml{T}, \fml{S}\cup \fml{T})$}
       \Else
           \Comment{Fix one feature}
           \State{$j\gets\msf{PickRandom}(\{ i\in\fml{T} \mid \oper{AE}_i = \True \})$}
           \State\Return{($\fml{W}\setminus \{j\},\fml{S})$}
       \EndIf
\EndProcedure
\end{algorithmic}

%% file: xps.tex
\section{Explanations on Demand}
\label{sec:xps}

\begin{algorithm}[ht]
  \input{./algs/allxp}
  \caption{MARCO enumeration of \teaxp/\tecxp}
  \label{alg:xp:all}
\end{algorithm}

\subsection{{\tecxp}s/{\teaxp}s Enumeration}
Besides computing one {\tecxp}/{\teaxp}, one may be
interested in navigating the sets of {\tecxp}s/{\teaxp}s 
(i.e.\ $\dsym\mbb{C}/\dsym\mbb{A}$). For example, 
we may be interested in deciding whether a
sensitive feature can occur in some explanation, or 
aggregate explanations of $\dsym\mbb{C}$/$\dsym\mbb{A}$ to 
compute \emph{feature importance score}~\citep{izza-aaai24b,Ignatiev-sat24b,lhams-corr24a}.
\cref{alg:xp:all} details a MARCO-like~\citep{lpmms-cj16} approach for
enumerating both all AXps and CXps. This algorithm illustrates the
integration of Boolean Satisfiability (SAT) algorithm with robustness 
reasoners, with the purpose of computing {\tecxp}s/{\teaxp}s.

\input{scxp}

%% file: algs/allxp.tex
\begin{flushleft}
  \hspace*{\algorithmicindent}
  \textbf{Input}: {
    Argument $\epsilon$,
    Parameters $\fml{E}$, $p$
  } 
  \hspace*{\algorithmicindent}
\end{flushleft}

\begin{algorithmic}[1]
  \State{$\fml{H}\gets\emptyset$}%
  \Comment{$\fml{H}$ defined on set $U=\{u_1,\ldots,u_m\}$}
  \Repeat 
  \State{$(\msf{outc},\nu)\gets\SAT(\fml{H})$}
  \If{$\msf{outc}=\TRUE$}
  \State{$\fml{X}\gets\{i\in\fml{F}\,|\,\nu(i)=0\}$}%
  \Comment{Fixed features}
  \State{$\fml{Y}\gets\{i\in\fml{F}\,|\,\nu(i)=1\}$}%
  \Comment{Free features} 
  \If{$\wcxp(\fml{Y},\epsilon;\fml{E})$}
  \Comment{$\fml{Y}\supseteq$ some \tecxp}
  \State{$\fml{S}\gets\findcxp(\epsilon,\fml{Y};\fml{E})$}
  \State{$\prtcxp(\fml{S})$}
  \State{$\fml{H}\gets\fml{H}\cup\{(\lor_{i\in\fml{S}}\neg{u_i})\}$}
  \Else
  \Comment{$\fml{X}\supseteq$ some \teaxp}
  \State{$\fml{S}\gets\findaxp(\epsilon,\fml{X};\fml{E},p)$}
  \State{$\prtaxp(\fml{S})$}
  \State{$\fml{H}\gets\fml{H}\cup\{(\lor_{i\in\fml{S}}{u_i})\}$}
  \EndIf
  \EndIf
  \Until{$\msf{outc}=\FALSE$}
\end{algorithmic}

%% file: scxp.tex
\subsection{Computing one Smallest \tecxp}

\paragraph{A MaxSMT/MaxSAT Formulation.}
Computing smallest (or minimum-size) \tecxp can be viewd 
as maximizing a solution in constrained optimization 
problems (MaxSAT, MaxSMT, etc).
Concretely, the problem is formulated as follow. 
We associate Boolean variables $s_i$ for $i\in\fml{F}$ s.t. 
$s_i=1$ iff feature $i\in\fml{F}$ is fixed 
 (i.e. $s_i\leftrightarrow(x_i=v_i)$). Thus, if
  $s_i=0$, then the value of $x_i$ needs not be equal to $v_i$.
Then, we define a set of hard constraints  $\fml{B}$ and 
soft $\fml{S}$ to satisfy, where, 
\begin{align*}  
 \fml{B} ~ = ~ 
  (\kappa(\mbf{x})=c)\land &
  (||(\mbf{x}-\mbf{v})||_{l_p}\le\epsilon)\land \\
  & \left[\bigwedge\nolimits_{i\in{\fml{F}}}s_i\leftrightarrow(x_i=v_i)\right]
\end{align*} 
and $\fml{S} = \{ (s_1), (s_2),\ldots, (s_m) \}$. Note that this solution 
is suitable for models that enable logical encoding.

\begin{claim}
  The MaxSMT/MaxSAT solution of $(\fml{B},\fml{S})$ is a smallest CXp.
\end{claim}

\begin{remark}
  We can solve the MaxSMT/MaxSAT formulation with an off-the-shelf
  MaxSMT/MaxSAT solver.
\end{remark}

\begin{algorithm}[ht]
  \input{./algs/smallcxp}
  \caption{Abstraction refinement for min \tecxp}
  \label{alg:scxp}
\end{algorithm}

\paragraph{An Abstraction Refinement Approach.} 
Let $\fml{H}$ represent an under-approximation of 
$\dsym\mbb{A}$, defined on a set of variables $u_i$, where $u_i=1$ denotes that
the picked set contains feature $i\in\fml{F}$, and so feature $i$ is
fixed in that set. 
(Pseudo-code of the approach is depicted in~\cref{alg:scxp}.)
Basically, the idea is to have a minimum model of $\fml{H}$, i.e.\ a minimum
hitting set of the (approximation) of $\dsym\mbb{A}$. If such minimum
hitting set corresponds to a \tewcxp, then it must be a minimum-size \tecxp.
Note that \cref{alg:scxp} can be easily adapted to find smallest {\teaxp}, i.e.\
minimum hitting set of $\dsym\mbb{C}$.
One key observation is that, for efficiency reasons, highly sophisticated 
NP solvers (off-the-shelf MaxSAT solvers)     
are preferred over abstract refinement algorithm for searching optimal solution 
of the contrastive explanation problem; whilst 
for smallest abductive explanation,  abstract refinement is often applied.

%% file: algs/smallcxp.tex
\begin{flushleft}
  \hspace*{\algorithmicindent}
  \textbf{Input}: {
    Argument $\epsilon$,
    Parameters $\fml{E}$, $p$
  } \\
  \hspace*{\algorithmicindent}
  \textbf{Output}: {Smallest \tecxp $\fml{S}$}  
\end{flushleft}

\begin{algorithmic}[1]
  \State{$\fml{H}\gets\emptyset$}%
  \Comment{$\fml{H}$ defined on set $U=\{u_1,\ldots,u_m\}$}
  \Repeat 
  \State{$(\msf{outc},\nu)=\msf{MinimumModel}(\fml{H})$}
  \If{$\msf{outc}=\TRUE$}
  \State{$\fml{S}=\{i\,|\,\nu(i)=1\}$}%
  \If{$\neg\wcxp(\fml{Y},\epsilon;\fml{E})$}
  \Comment{$\fml{S}\supseteq$ \teaxp}
  \State{$\fml{A}\gets\findaxp(\epsilon,\fml{F}\setminus\fml{S};\fml{E},p)$}
  \State{$\fml{H}\gets\fml{H}\cup\{(\lor_{i\in\fml{A}}{u_i})\}$}
  \EndIf
  \EndIf
  \Until{$\msf{outc}=\FALSE$}
  \State{\tbf{return} $\fml{S}$}
  \Comment{minimum-size \tecxp}
\end{algorithmic}

%% file: res.tex
\section{Experiments} \label{sec:res}
We assess our approach \swiftcxp to computing \tecxp for DNNs 
on well known image data. 
Additionally, we analyze feature/pixel importance scores 
by applying (partial) enumeration of {\tecxp}s.

\input{./figs/cxp}

\subsection{Evaluation Setting}
\paragraph{Experimental Setup.}
All experiments were carried out on a high-performance computer
cluster with machines equipped with AMD EPYC 7713 processors.
Each instance test is provided with 2 and 60 cores, resp., when running
the  dichotomic (\cref{alg:dicho})
and our parallel \swiftcxp algorithm, namely 1 core for 1 oracle used
and 1 additional core to run the main script.
Furthermore, the memory limit was set to 16GB, and the time limit
to 14400 seconds (i.e.\ 4 hours).

\paragraph{Prototype Implementation.}
The proposed approach was prototyped as a set of Python scripts\footnote{%
Code will be released at \url{https://github.com/izzayacine/SwiftXPlain} 
after acceptance of the paper.}, %
and PyTorch library \citep{pytorch-nips19} was used to  train and
handle the learned DNNs.
A unified Python interface for  robustness oracles is implemented and it
enables us to use any DNN reasoner of the VNN-COM~\citep{johnson-sttt23}.
MN-BaB~\citep{Vechev-iclr22}, which is a complete neural network
verifier, is used to instrument AEx checking on CPU mode. 
Moreover, Gurobi~\citep{gurobi} MILP solver is applied for empowering 
MN-BaB resolution.
Besides, we implemented  the {\it pixel sensitivity} ranking heuristic
proposed in~\citep{barrett-nips23} for the traversal order of features 
in all algorithms outlined above.
Note that we also tested LIME~\citep{guestrin-kdd16} as another 
heuristic but performs poorly\footnote{%
There are many features highly ranked by LIME but  are 
not included in the explanation; conversely,  features that 
are assigned lower scores but are relevant for the explanation.}  
compared to pixel sensitivity.
Additionally, we implement a heuristic to approximate $\fml{W}$. Concretely,  
we utilize the returned $\mbf{u}$ \taex at the initial oracle call in the algorithm 
and to construct a binary mask ($\lVert\mbf{u}-\mbf{v}\rVert > 0$)  and apply 
on vector $\mbf{x}$ of (ordered) feature input, thus $\mbf{x}$ becomes 
sparse after masking out pixels $i$ where $|u_i - v_i| = 0$ and 
$\fml{W}$ represents the set of unmasked pixels. 
\paragraph{Image Recognition Benchmarks.}
The experiments focus on two well-known image datasets, that
have been studied in~\citep{barrett-nips23} for \teaxp.
Namely, we evaluate the widely used {MNIST}~\citep{Li-spm12}
dataset, which features hand-written digits from 0 to 9.
Also, we consider the image dataset {GTSRB}~\citep{StallkampSSI12} of traffic signs, and
we select a collection of training data that represents the top 10 classes of
the entire data.
We considered different $\epsilon$ values for each model and image size, 
e.g. for {MNIST} benchmark
$\epsilon$  varies from $0.08$ to $0.15$.
The parameter $\delta$ in \swiftcxp, to activate feature 
disjunction procedure, is varied from $0.75$ 
to $0.9$ w.r.t. the distance $\epsilon$ and the image size.

\subsection{Results}
\paragraph{Computing One \tecxp.}
\cref{tab:cxp} summarizes the results comparing \swiftcxp and
the baseline dichotomic method. on fully-connected (dense)
and convolutional NNs  trained with the above image datasets.
As can be observed from \cref{tab:cxp}, \swiftcxp
significantly outperforms the dichotomic search on all tested
benchmarks.
More importantly, the dichotomic approach is unable to deliver a \tecxp,  
within 4-hours,
on the majority of benchmarks, e.g.\  20\% and 35\% for 
small {\it mnist} NNs, and fails on all tests on
large (convolutional) models. 
In contrast, \swiftcxp successfully finds an explanation on all
tests with an avg. runtime of $932.8$ sec ($\sim$16 min) on 
the largest NN {\it mnist-conv}.
%
%
%
%
Focusing solely on the performance of \swiftcxp, we observe that
activation of the feature disjunction (FD) technique is more effective
when the feature set size $|\fml{W}|$ left to inspect is smaller than
the average size of \teaxp.
Also, we observe that deactivating FD in the first iterations of
\swiftcxp enables us to drop  up to 70\% of (chunks of) features with a
few iterations in the inner loop. 
Furthermore, one can see from \cref{tab:cxp} that the average
success of FD to capture $q$ transition features in one iteration
varies from $20\%$ to $100\%$ for {MNIST} and $100\%$ 
for {GTSRB}.
%
%
Moreover, we note that sensitivity feature traversal 
strategy improves the effectiveness of FD --- positioning 
all relevant pixels at the bottom of $\fml{W}$ increases 
the number of successful FD (parallel) calls.
Regarding approximation heuristic, we observe a significant 
gain on {\it mnist} NNs --- up to 60\% pixels $\fml{F}$ are 
discarded, however poorly performs on {\it gtsrb}; hence 
in future work, we are willing to analyze the robustness reasoner  
to optimize the \taex bounds.

To conclude, observations above clearly show that \cref{alg:fd} and
\cref{alg:swiftxp} synergizes to analyze (free/fix) chunks of feature 
in parallel, s.t. when the input
data or a target \tecxp is expected to be large 
then FD serves to augment $\fml{S}$ by means
of fewer iterations; conversely for larger $\epsilon$ distance or
when a smaller \tecxp is expected, \cref{alg:swiftxp} allows us to
eliminate more features with fewer iterations.

\paragraph{Enumerating {\tecxp}s \& Feature Importance.}
We run additional experiments for enumerating explanations for two NNs benchmarks 
{\it gtrb-convSmall} and {\it mnist-dense}, and the number of {\tecxp}s to enumerate 
per image sample is fixed to 100. 
Moreover, we compute feature attribution/importance score 
(FFA)~\cite{izza-aaai24b,Ignatiev-sat24b,ignatiev-corr23a,lhams-corr24a} 
to generate feature attribution explanations, such that
$$ 
 i\in\fml{F},\:\: 
 \FFA(i) = \frac{|\{\fml{S} \mid \fml{S}\in\dsym\mbb{C}(\fml{E},\refd;p), i\in\fml{S} \}|}{\dsym\mbb{C}(\fml{E},\refd;p)}  
$$
and feature attribution \tecxp denoted by FFA-\tecxp is  
the set of features $i$ 
for which $\FFA(i) > 0$.

\cref{fig:img} showcases visual interpretation of computed distance-restricted 
contrastive explanations  
for 2 image examples --- speed limit image for GTSRB and digit `5' image  
for MNIST.   
One can observe that the generated {\tecxp}s for the selected samples are sparse,
i.e.\ small w.r.t. image size.     
particularly,  for GTSRB sample the average 
length of enumerated \tecxp in $\dsym\mbb{C}$ is $59.27$ (pixels) and 
160 for FFA-\tecxp. Moreover, the average runtime to compute enumerate 
explanations of  $\dsym\mbb{C}$ is 253 seconds for GTSRB and 
7133 seconds  for MNIST. 

\input{imgXP}

Furthermore, we observe that the distribution of features/pixels involved 
in the set of explanations $\dsym\mbb{C}$  is gaussian, namely  {\tecxp}s 
share often the same features, which yields a succinct feature 
attribution \tecxp and highly scored relevant pixels.

%% file: figs/cxp.tex
\sisetup{parse-numbers=false,detect-all,mode=text}
\setlength{\tabcolsep}{4pt}

\begin{table*}[ht]
\centering
\resizebox{\textwidth}{!}{
  \begin{tabular}{lS[table-format=1.3]S[table-format=2]
  S[table-format=1.2]S[table-format=4]
  S[table-format=3]S[table-format=3.1]S[table-format=3.1]S[table-format=3.1]S[table-format=3]
  S[table-format=2.2]S[table-format=3]S[table-format=3] 
  S[table-format=2.1]S[table-format=4.1]S[table-format=4.1] 
  >{\columncolor{platinum!80}} S[table-format=4.1]}
\toprule[1.2pt]
\multirow{2}{*}{\bf Model}  & \multirow{2}{*}{$\epsilon_\infty$} & {\bf \tewcxp} &
\multicolumn{7}{c}{\bf Dichotomic search} & \multicolumn{7}{c}{\bf  \swiftcxp}  \\
  \cmidrule[0.8pt](lr{.75em}){3-3}
  \cmidrule[0.8pt](lr{.75em}){4-10}
  \cmidrule[0.8pt](lr{.75em}){11-17}
\rowcolor{white} 
&  & {\bf Len\%} & {\bf avgC}  & {\bf nCalls} & {\bf Len} & {\bf Mn}  & {\bf Mx} & {\bf avg } &  {\bf TO\%} &
		{\bf avgC}  &  {\bf nCalls} & {\bf Len} & {\bf FD\%} & {\bf Mn}  & {\bf Mx} & {\bf avg}  \\
\toprule[1.2pt]

gtsrb-dense &  0.15 &  99 & $\textemdash$ & $\textemdash$ & $\textemdash$ & $\textemdash$ & $\textemdash$ & $\textemdash$ & 100 & 1.45 & 34 & 216 & \cellcolor{midblue!25} 100 & 53.6 & 94.3 &  68.4 \\
gtsrb-convSmall &  0.008 &  99 & $\textemdash$ & $\textemdash$ & $\textemdash$ & $\textemdash$ & $\textemdash$ & $\textemdash$ & 100 & 0.28 & 61 & 130 & \cellcolor{midblue!25} 100 & 15.1 & 21.7 &  17.9 \\
mnist-dense &  0.08 &  59 & 0.49 & 1235 & 186 & 593.3 & 633.3 & 610.7 & 20 & 6.27 & 198 & 186 & 25.4 & 165.1 & 895.9 &  588.1 \\
mnist-denseSmall &  0.08 &  49 & 0.19 & 386 & 67 & 72.4 & 77.3 & 75.1 & 35 & 2.10 & 97 & 66 & 15.1 & 53.4 & 770.2 &  91.3 \\
mnist-conv &  0.15 &  72 & $\textemdash$ & $\textemdash$ & $\textemdash$ & $\textemdash$ & $\textemdash$ & $\textemdash$ & 100 & 19.43 & 48 & 572 & \cellcolor{midblue!25} 100 & 693.3 & 1185.2 &  932.8 \\

\bottomrule[1.2pt]
\end{tabular}
}
\caption{%
\footnotesize{
  Detailed performance evaluation of computing \teaxp for DNNs with 
  \swiftcxp and comparison with the baseline dichotomic search algorithm. 
The number of processes in \swiftcxp is fixed to 30 (CPUs) for all tested models.
%
%
%
Columns  {\bf avgC} and  {\bf nCalls}  report, resp.\, the average time 
and average number of instrumented (AEx robustness) oracle  calls.
Column {\bf avg} (resp.\ {\bf Mn} and {\bf Mx})  reports the average  
(resp.\ min and max) time in seconds to deliver a \teaxp, and 
{\bf TO} is the percentage of timeout tests. 
Lastly, column {\bf Len} reports the average explanation 
length and {\bf FD\%} is the average percentage of successful FD 
calls to augment $\fml{S}$ with $q$ features in one iteration. 
 } }
\label{tab:cxp}
\vspace*{-0.3cm}
\end{table*}

%% file: imgXP.tex
\begin{figure*}[ht]
   \centering
   \begin{subfigure}[b]{0.25\textwidth}
   	\centering
       \includegraphics[scale=0.3]{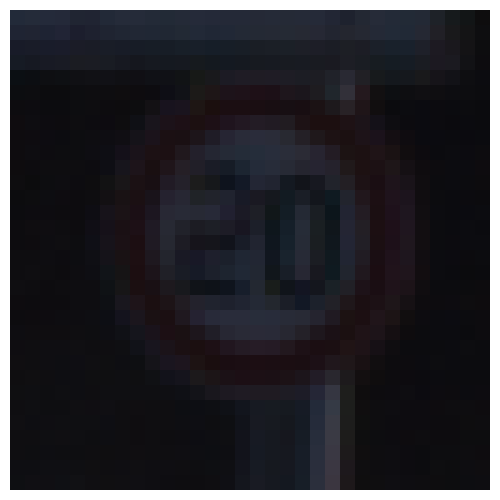}
       \captionsetup{width=0.75\textwidth}
       \caption{Original GTSRB traffic sign image} 
    \end{subfigure}
   \begin{subfigure}[b]{0.25\textwidth}
   	\centering
	\captionsetup{width=0.75\textwidth}
       \includegraphics[scale=0.3]{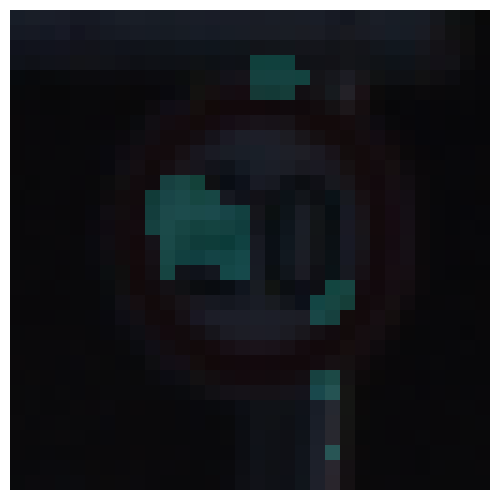} 
       \caption{Highlighted explanation pixels s.t. $\dsym=0.008$ } 
    \end{subfigure}
   \begin{subfigure}[b]{0.25\textwidth}
   	\centering
	\captionsetup{width=0.75\textwidth}
       \includegraphics[scale=0.3]{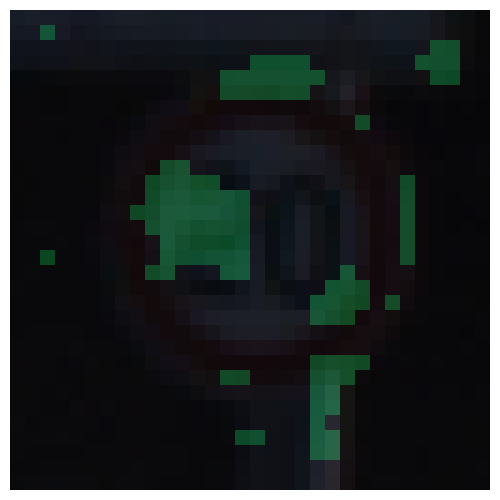}
       \caption{Highlighted FFA score pixels s.t. $\dsym=0.008$}  
    \end{subfigure}        
    \vskip\baselineskip   
   \begin{subfigure}[b]{0.25\textwidth}
   	\centering
	\captionsetup{width=0.75\textwidth}
       \includegraphics[scale=0.3]{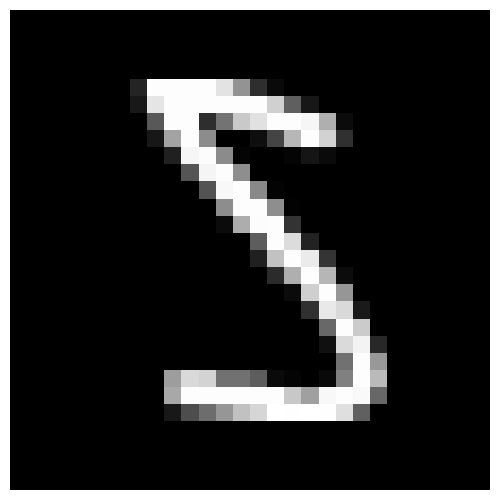}
       \caption{Original MNIST digit image} 
    \end{subfigure}
   \begin{subfigure}[b]{0.25\textwidth}
   	\centering
	\captionsetup{width=0.75\textwidth}
       \includegraphics[scale=0.3]{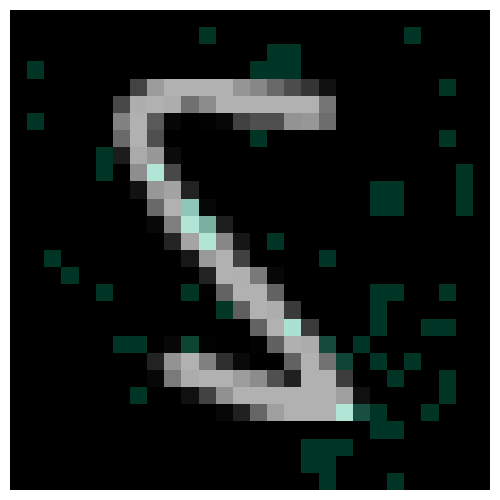} 
       \caption{Highlighted explanation pixels s.t. $\dsym=0.08$ } 
    \end{subfigure}
   \begin{subfigure}[b]{0.25\textwidth}
   	\centering
	\captionsetup{width=0.75\textwidth}
       \includegraphics[scale=0.3]{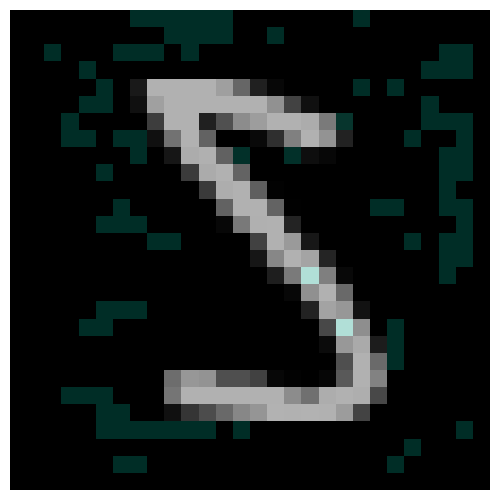}
       \caption{Highlighted FFA score pixels s.t. $\dsym=0.08$}  
    \end{subfigure}       
   \caption{Visualize \emph{Feature Attribution}-based Contrastive explanations for image datasets: MNIST and GTSRB.}
   \label{fig:img}
\end{figure*}

%% file: conc.tex
\section{Conclusions}
\label{sec:conc}

The importance of computing rigorous explanations cannot be
overstated. Recent work proposed a novel approach for computing
sufficient reasons for a prediction, by restricting the validity of
explanations to within a maximum distance of the
sample~\citep{barrett-nips23,barrett-corr24,hms-corr23,swiftxp-kr24}, i.e.\ the so-called distance-restricted
explanations.
This paper extends this earlier work in several directions, proposing
novel algorithms for computing distance-restricted contrastive
explanations, and the deterministic enumeration of both contrastive
and abductive explanations, which serves to compute feature 
importance score for both type of explanations. Furthermore, the paper 
also investigates the computation of smallest (minimum cardinality) 
distance-restricted contrastive explanations.
%

%% file: replbib.tex
\newtoggle{mkbbl}

%% file: togbbl.tex
\settoggle{mkbbl}{false}

%% file: paper.bbl
\begin{thebibliography}{}

\bibitem[\protect\citeauthoryear{ACM}{2019}]{acm-turing-award-2018}
ACM.
\newblock Fathers of the deep learning revolution receive {ACM} {A.~M.~T}uring
  award.
\newblock \url{https://awards.acm.org/about/2018-turing}, 2019.

\bibitem[\protect\citeauthoryear{BBC}{2024a}]{bbc-turing-award-2018}
BBC.
\newblock British-{C}anadian {AI} expert {G}eoffrey {H}inton wins {T}uring
  award.
\newblock \url{https://www.bbc.com/news/technology-47721129}, 2024.

\bibitem[\protect\citeauthoryear{BBC}{2024b}]{bbc-physics-nobel}
BBC.
\newblock 'godfather of ai' shares {N}obel physics prize.
\newblock \url{https://www.bbc.com/news/articles/c62r02z75jyo}, 2024.

\bibitem[\protect\citeauthoryear{BBC}{2024c}]{bbc-chemistry-nobel}
BBC.
\newblock Google {D}eep{M}ind boss wins {N}obel for proteins breakthrough.
\newblock \url{https://www.bbc.com/news/articles/czrm0p2mxvyo}, 2024.

\bibitem[\protect\citeauthoryear{Biere \bgroup \em et al.\egroup
  }{2021}]{sat-handbook21}
Armin Biere, Marijn Heule, Hans van Maaren, and Toby Walsh, editors.
\newblock {\em Handbook of Satisfiability - Second Edition}.
\newblock 2021.

\bibitem[\protect\citeauthoryear{Biradar \bgroup \em et al.\egroup
  }{2024}]{izza-aaai24b}
Gagan Biradar, Yacine Izza, Elita~A. Lobo, Vignesh Viswanathan, and Yair Zick.
\newblock Axiomatic aggregations of abductive explanations.
\newblock In {\em AAAI}, pages 11096--11104, 2024.

\bibitem[\protect\citeauthoryear{Brix \bgroup \em et al.\egroup
  }{2023}]{johnson-sttt23}
Christopher Brix, Mark~Niklas M{\"{u}}ller, Stanley Bak, Taylor~T. Johnson, and
  Changliu Liu.
\newblock First three years of the international verification of neural
  networks competition {(VNN-COMP)}.
\newblock {\em Int. J. Softw. Tools Technol. Transf.}, 25(3):329--339, 2023.

\bibitem[\protect\citeauthoryear{Darwiche}{2023}]{darwiche-lics23}
Adnan Darwiche.
\newblock Logic for explainable {AI}.
\newblock In {\em {LICS}}, pages 1--11, 2023.

\bibitem[\protect\citeauthoryear{Deng}{2012}]{Li-spm12}
Li~Deng.
\newblock The mnist database of handwritten digit images for machine learning
  research.
\newblock {\em IEEE Signal Processing Magazine}, 29(6):141--142, 2012.

\bibitem[\protect\citeauthoryear{Ferrari \bgroup \em et al.\egroup
  }{2022}]{Vechev-iclr22}
Claudio Ferrari, Mark~Niklas M{\"{u}}ller, Nikola Jovanovic, and Martin~T.
  Vechev.
\newblock Complete verification via multi-neuron relaxation guided
  branch-and-bound.
\newblock In {\em ICLR}, 2022.

\bibitem[\protect\citeauthoryear{{Gurobi Optimization, LLC}}{2023}]{gurobi}
{Gurobi Optimization, LLC}.
\newblock {Gurobi Optimizer Reference Manual}, 2023.

\bibitem[\protect\citeauthoryear{Hemery \bgroup \em et al.\egroup
  }{2006}]{sais-ecai06}
Fred Hemery, Christophe Lecoutre, Lakhdar Sais, and Fr{\'{e}}d{\'{e}}ric
  Boussemart.
\newblock Extracting {MUCs} from constraint networks.
\newblock In {\em ECAI}, pages 113--117, 2006.

\bibitem[\protect\citeauthoryear{Horn and Johnson}{2012}]{horn-bk12}
Roger~A. Horn and Charles~R. Johnson.
\newblock {\em Matrix Analysis, 2nd Ed}.
\newblock Cambridge University Press, 2012.

\bibitem[\protect\citeauthoryear{Huang and Marques{-}Silva}{2023a}]{hms-ecai23}
Xuanxiang Huang and Jo{\~{a}}o Marques{-}Silva.
\newblock From decision trees to explained decision sets.
\newblock In {\em ECAI}, pages 1100--1108, 2023.

\bibitem[\protect\citeauthoryear{Huang and Marques{-}Silva}{2023b}]{hms-corr23}
Xuanxiang Huang and Jo{\~{a}}o Marques{-}Silva.
\newblock From robustness to explainability and back again.
\newblock {\em CoRR}, abs/2306.03048, 2023.

\bibitem[\protect\citeauthoryear{Huang \bgroup \em et al.\egroup
  }{2021}]{hiims-kr21}
Xuanxiang Huang, Yacine Izza, Alexey Ignatiev, and Jo{\~{a}}o Marques{-}Silva.
\newblock On efficiently explaining graph-based classifiers.
\newblock In {\em KR}, pages 356--367, 2021.

\bibitem[\protect\citeauthoryear{Ignatiev and
  Marques{-}Silva}{2021}]{ims-sat21}
Alexey Ignatiev and Joao Marques{-}Silva.
\newblock {SAT}-based rigorous explanations for decision lists.
\newblock In {\em SAT}, pages 251--269, 2021.

\bibitem[\protect\citeauthoryear{Ignatiev \bgroup \em et al.\egroup
  }{2019}]{inms-aaai19}
Alexey Ignatiev, Nina Narodytska, and Joao Marques{-}Silva.
\newblock Abduction-based explanations for machine learning models.
\newblock In {\em AAAI}, pages 1511--1519, 2019.

\bibitem[\protect\citeauthoryear{Ignatiev \bgroup \em et al.\egroup
  }{2020}]{inams-aiia20}
Alexey Ignatiev, Nina Narodytska, Nicholas Asher, and Joao Marques{-}Silva.
\newblock From contrastive to abductive explanations and back again.
\newblock In {\em AIxIA}, pages 335--355, 2020.

\bibitem[\protect\citeauthoryear{Ignatiev \bgroup \em et al.\egroup
  }{2021}]{Ignatiev-aaai21}
Alexey Ignatiev, Edward Lam, Peter~J. Stuckey, and Jo{\~{a}}o Marques{-}Silva.
\newblock A scalable two stage approach to computing optimal decision sets.
\newblock In {\em AAAI}, pages 3806--3814, 2021.

\bibitem[\protect\citeauthoryear{Ignatiev \bgroup \em et al.\egroup
  }{2022}]{iisms-aaai22}
Alexey Ignatiev, Yacine Izza, Peter~J. Stuckey, and Jo{\~{a}}o Marques{-}Silva.
\newblock Using {MaxSAT} for efficient explanations of tree ensembles.
\newblock In {\em AAAI}, pages 3776--3785, 2022.

\bibitem[\protect\citeauthoryear{Izza and Marques{-}Silva}{2021}]{ims-ijcai21}
Yacine Izza and Joao Marques{-}Silva.
\newblock On explaining random forests with {SAT}.
\newblock In {\em IJCAI}, pages 2584--2591, 2021.

\bibitem[\protect\citeauthoryear{Izza and
  Marques{-}Silva}{2023}]{ims-workshop-kr23}
Yacine Izza and Jo{\~{a}}o Marques{-}Silva.
\newblock On computing relevant features for explaining nbcs.
\newblock pages 75--86, 2023.

\bibitem[\protect\citeauthoryear{Izza \bgroup \em et al.\egroup
  }{2020}]{iims-corr20}
Yacine Izza, Alexey Ignatiev, and Joao Marques{-}Silva.
\newblock On explaining decision trees.
\newblock {\em CoRR}, abs/2010.11034, 2020.

\bibitem[\protect\citeauthoryear{Izza \bgroup \em et al.\egroup
  }{2022a}]{iims-jair22}
Yacine Izza, Alexey Ignatiev, and Jo{\~{a}}o Marques{-}Silva.
\newblock On tackling explanation redundancy in decision trees.
\newblock {\em J. Artif. Intell. Res.}, 75:261--321, 2022.

\bibitem[\protect\citeauthoryear{Izza \bgroup \em et al.\egroup
  }{2022b}]{iincms-corr22}
Yacine Izza, Alexey Ignatiev, Nina Narodytska, Martin~C. Cooper, and Jo{\~{a}}o
  Marques{-}Silva.
\newblock Provably precise, succinct and efficient explanations for decision
  trees.
\newblock {\em CoRR}, abs/2205.09569, 2022.

\bibitem[\protect\citeauthoryear{Izza \bgroup \em et al.\egroup
  }{2023}]{ihincms-ijar23}
Yacine Izza, Xuanxiang Huang, Alexey Ignatiev, Nina Narodytska, Martin~C.
  Cooper, and Jo{\~{a}}o Marques{-}Silva.
\newblock On computing probabilistic abductive explanations.
\newblock {\em Int. J. Approx. Reason.}, 159:108939, 2023.

\bibitem[\protect\citeauthoryear{Izza \bgroup \em et al.\egroup
  }{2024a}]{swiftxp-kr24}
Yacine Izza, Xuanxiang Huang, Antonio Morgado, Jordi Planes, Alexey Ignatiev,
  and Joao Marques-Silva.
\newblock {Distance-Restricted Explanations: Theoretical Underpinnings \&
  Efficient Implementation}.
\newblock In {\em KR}, pages 475--486, 2024.

\bibitem[\protect\citeauthoryear{Izza \bgroup \em et al.\egroup
  }{2024b}]{iisms-aaai24}
Yacine Izza, Alexey Ignatiev, Peter~J. Stuckey, and Jo{\~{a}}o Marques{-}Silva.
\newblock Delivering inflated explanations.
\newblock In {\em AAAI}, pages 12744--12753, 2024.

\bibitem[\protect\citeauthoryear{Izza \bgroup \em et al.\egroup
  }{2024c}]{izza-ecai24}
Yacine Izza, Kuldeep~S. Meel, and Jo{\~{a}}o Marques{-}Silva.
\newblock Locally-minimal probabilistic explanations.
\newblock In {\em ECAI}, volume 392, pages 1092--1099, 2024.

\bibitem[\protect\citeauthoryear{Letoffe \bgroup \em et al.\egroup
  }{2024a}]{lhams-corr24a}
Olivier Letoffe, Xuanxiang Huang, Nicholas Asher, and Jo{\~{a}}o
  Marques{-}Silva.
\newblock From {SHAP} scores to feature importance scores.
\newblock {\em CoRR}, abs/2405.11766, 2024.

\bibitem[\protect\citeauthoryear{Letoffe \bgroup \em et al.\egroup
  }{2024b}]{lhms-aaai25}
Olivier Letoffe, Xuanxiang Huang, and Jo{\~{a}}o Marques{-}Silva.
\newblock On correcting {SHAP} scores.
\newblock In {\em AAAI}, 2024.

\bibitem[\protect\citeauthoryear{Liffiton \bgroup \em et al.\egroup
  }{2016}]{lpmms-cj16}
Mark~H. Liffiton, Alessandro Previti, Ammar Malik, and Jo{\~{a}}o
  Marques{-}Silva.
\newblock Fast, flexible {MUS} enumeration.
\newblock {\em Constraints An Int. J.}, 21(2):223--250, 2016.

\bibitem[\protect\citeauthoryear{Marques{-}Silva and
  Menc{\'{\i}}a}{2020}]{msm-ijcai20}
Jo{\~{a}}o Marques{-}Silva and Carlos Menc{\'{\i}}a.
\newblock Reasoning about inconsistent formulas.
\newblock In {\em IJCAI}, pages 4899--4906, 2020.

\bibitem[\protect\citeauthoryear{Marques{-}Silva \bgroup \em et al.\egroup
  }{2013a}]{mshjpb-ijcai13}
Jo{\~{a}}o Marques{-}Silva, Federico Heras, Mikol{\'{a}}s Janota, Alessandro
  Previti, and Anton Belov.
\newblock On computing minimal correction subsets.
\newblock In {\em IJCAI}, pages 615--622, 2013.

\bibitem[\protect\citeauthoryear{Marques{-}Silva \bgroup \em et al.\egroup
  }{2013b}]{msjb-cav13}
Jo{\~{a}}o Marques{-}Silva, Mikol{\'{a}}s Janota, and Anton Belov.
\newblock Minimal sets over monotone predicates in boolean formulae.
\newblock In {\em CAV}, pages 592--607, 2013.

\bibitem[\protect\citeauthoryear{Marques{-}Silva \bgroup \em et al.\egroup
  }{2017}]{msjm-aij17}
Jo{\~{a}}o Marques{-}Silva, Mikol{\'{a}}s Janota, and Carlos Menc{\'{\i}}a.
\newblock Minimal sets on propositional formulae. problems and reductions.
\newblock {\em Artif. Intell.}, 252:22--50, 2017.

\bibitem[\protect\citeauthoryear{Marques{-}Silva \bgroup \em et al.\egroup
  }{2020}]{NBC-nips20}
Jo{\~{a}}o Marques{-}Silva, Thomas Gerspacher, Martin~C. Cooper, Alexey
  Ignatiev, and Nina Narodytska.
\newblock Explaining naive bayes and other linear classifiers with polynomial
  time and delay.
\newblock In {\em NeurIPS}, 2020.

\bibitem[\protect\citeauthoryear{Marques{-}Silva}{2022}]{ms-rw22}
Jo{\~{a}}o Marques{-}Silva.
\newblock Logic-based explainability in machine learning.
\newblock In {\em Reasoning Web}, pages 24--104, 2022.

\bibitem[\protect\citeauthoryear{Marques{-}Silva}{2024}]{ms-isola24}
Jo{\~{a}}o Marques{-}Silva.
\newblock Logic-based explainability: Past, present and future.
\newblock In {\em ISoLA}, pages 181--204, 2024.

\bibitem[\protect\citeauthoryear{Miller}{2019}]{miller-aij19}
Tim Miller.
\newblock Explanation in artificial intelligence: Insights from the social
  sciences.
\newblock {\em Artif. Intell.}, 267:1--38, 2019.

\bibitem[\protect\citeauthoryear{Paszke \bgroup \em et al.\egroup
  }{2019}]{pytorch-nips19}
Adam Paszke, Sam Gross, Francisco Massa, Adam Lerer, James Bradbury, Gregory
  Chanan, Trevor Killeen, Zeming Lin, Natalia Gimelshein, Luca Antiga, Alban
  Desmaison, Andreas K{\"{o}}pf, Edward~Z. Yang, Zachary DeVito, Martin Raison,
  Alykhan Tejani, Sasank Chilamkurthy, Benoit Steiner, Lu~Fang, Junjie Bai, and
  Soumith Chintala.
\newblock Pytorch: An imperative style, high-performance deep learning library.
\newblock In {\em NeurIPS}, pages 8024--8035, 2019.

\bibitem[\protect\citeauthoryear{Ribeiro \bgroup \em et al.\egroup
  }{2016}]{guestrin-kdd16}
Marco~T{\'{u}}lio Ribeiro, Sameer Singh, and Carlos Guestrin.
\newblock "why should {I} trust you?": Explaining the predictions of any
  classifier.
\newblock In {\em KDD}, pages 1135--1144, 2016.

\bibitem[\protect\citeauthoryear{Robinson}{2003}]{robinson-bk03}
Derek~JS Robinson.
\newblock {\em An introduction to abstract algebra}.
\newblock Walter de Gruyter, 2003.

\bibitem[\protect\citeauthoryear{Shih \bgroup \em et al.\egroup
  }{2018}]{darwiche-ijcai18}
Andy Shih, Arthur Choi, and Adnan Darwiche.
\newblock A symbolic approach to explaining bayesian network classifiers.
\newblock In {\em IJCAI}, pages 5103--5111, 2018.

\bibitem[\protect\citeauthoryear{Stallkamp \bgroup \em et al.\egroup
  }{2012}]{StallkampSSI12}
Johannes Stallkamp, Marc Schlipsing, Jan Salmen, and Christian Igel.
\newblock Man vs. computer: Benchmarking machine learning algorithms for
  traffic sign recognition.
\newblock {\em Neural Networks}, 32:323--332, 2012.

\bibitem[\protect\citeauthoryear{Wu \bgroup \em et al.\egroup
  }{2022}]{barrett-corr22}
Min Wu, Haoze Wu, and Clark~W. Barrett.
\newblock {VeriX}: Towards verified explainability of deep neural networks.
\newblock {\em CoRR}, abs/2212.01051, 2022.

\bibitem[\protect\citeauthoryear{Wu \bgroup \em et al.\egroup
  }{2023}]{barrett-nips23}
Min Wu, Haoze Wu, and Clark~W. Barrett.
\newblock Verix: Towards verified explainability of deep neural networks.
\newblock In {\em NeurIPS}, 2023.

\bibitem[\protect\citeauthoryear{Wu \bgroup \em et al.\egroup
  }{2024}]{barrett-corr24}
Min Wu, Xiaofu Li, Haoze Wu, and Clark~W. Barrett.
\newblock Better verified explanations with applications to incorrectness and
  out-of-distribution detection.
\newblock {\em CoRR}, abs/2409.03060, 2024.

\bibitem[\protect\citeauthoryear{Yu \bgroup \em et al.\egroup
  }{2023}]{ignatiev-corr23a}
Jinqiang Yu, Alexey Ignatiev, and Peter~J. Stuckey.
\newblock On formal feature attribution and its approximation.
\newblock {\em CoRR}, abs/2307.03380, 2023.

\bibitem[\protect\citeauthoryear{Yu \bgroup \em et al.\egroup
  }{2024}]{Ignatiev-sat24b}
Jinqiang Yu, Graham Farr, Alexey Ignatiev, and Peter~J. Stuckey.
\newblock Anytime approximate formal feature attribution.
\newblock In {\em SAT}, pages 30:1--30:23, 2024.

\end{thebibliography}
